\documentclass[10pt,twocolumn,letterpaper]{article}

\usepackage{titling} 

\usepackage{3dv}
\usepackage{graphicx, graphics}
\usepackage{comment}
\usepackage{amsmath,amssymb} 
\usepackage{color}

\newcommand{\chk}{\checkmark}
\newcommand{\X}{$\times$}
\usepackage{enumitem}
\usepackage{bm}
\usepackage{subcaption}
\usepackage{varwidth}
\usepackage{mdframed}
\usepackage{epsfig}
\usepackage{booktabs}
\usepackage{multirow}
\usepackage{gensymb}
\usepackage{wrapfig}
\usepackage{placeins}

\usepackage[pagebackref=true,breaklinks=true,letterpaper=true,colorlinks,bookmarks=false]{hyperref}

\threedvfinalcopy 

\def\threedvPaperID{264} 
\def\httilde{\mbox{\tt\raisebox{-.5ex}{\symbol{126}}}}

\ifthreedvfinal\fi
\begin{document}

\title{Towards a MEMS-based Adaptive LIDAR}

\author{Francesco Pittaluga\thanks{Equal Contribution}\\
NEC Labs America\\
{\tt\small francescopittaluga@nec-labs.com}
\and
Zaid Tasneem$^{*}$\\
University of Florida\\
{\tt\small ztasneem@ufl.edu}
\and
Justin Folden$^{*}$\\
University of Florida\\
{\tt\small jfolden@ufl.edu}
\and
Brevin Tilmon\\
University of Florida\\
{\tt\small btilmon@.ufl.edu}
\and
Ayan Chakrabarti\\
Washington University in St. Louis\\
{\tt\small ayan@wustl.edu}
\and
Sanjeev J. Koppal\\
University of Florida\\
{\tt\small sjkoppal@ece.ufl.edu}
}

\maketitle

\begin{abstract}
We present a proof-of-concept LIDAR design that allows adaptive real-time measurements according to dynamically specified measurement patterns. We describe our optical setup and calibration, which enables fast sparse depth measurements using a scanning MEMS (micro-electro-mechanical) mirror. We validate the efficacy of our prototype LIDAR design by testing on over 75 static and dynamic scenes spanning a range of environments. We show CNN-based depth-map completion experiments which demonstrate that our sensor can realize adaptive depth sensing for dynamic scenes.

\end{abstract}

\section{Introduction}

Learning-enabled depth sensing has impacted every aspect of robotics. This success has prompted vision researchers to close the loop between active sensing and inference---with methods for correcting incomplete and imperfect depth measurements \cite{uhrig2017sparsity,zhang2018deep}, as well as those that decide where to sense next~\cite{liu2019neural,adaptivelidarstanford}.

However, such work is predicated on LIDAR systems that are flexible in how they make measurements. But this capability does not exist in most existing LIDAR hardware, where sampling is done in a set of fixed angles, usually modulated by slow mechanical motors.

\begin{figure*}[t]
\centering
    \includegraphics[width=\linewidth]{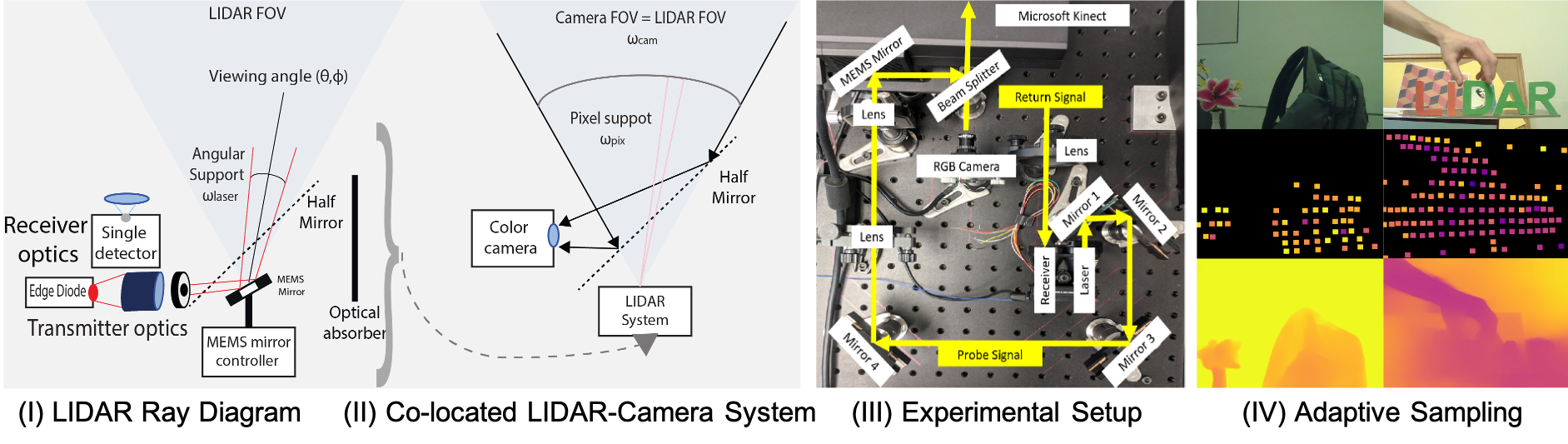}
    \caption{\textbf{Experimental setup.} We have designed a flexible MEMS mirror-modulated scanning LIDAR, as shown in (I). In (II), we co-locate this directionally controllable LIDAR with a color camera, allowing for deep depth completion of the sparse LIDAR measurements. In (III) we show a picture of the hardware setup corresponding to (I-II). The long optical path is simply an artifact of having a single circuit board for both the LIDAR receiver and transmitter. In (IV) we show adaptive sampling (middle) and deep depth completion (bottom) results captured with our Adaptive LIDAR Prototype.}
    \label{fig:setup}
\end{figure*}

We present a proof-of-concept, adaptive LIDAR platform that can leverage modern vision algorithms. It permits making measurements with different sampling patterns---providing a speed advantage when fewer measurements are made---and is co-located with a color camera to fully realize the benefits of deep depth completion and guided sampling. 

\subsection{Why Foveating LIDAR?}

Unlike most artificial sensors, animal eyes \emph{foveate}, or distribute resolution where it is needed. This is computationally efficient, since neuronal resources are concentrated on regions of interest. Similarly, we believe that an adaptive LIDAR would be useful on resource-constrained small platforms, such as micro-UAVs. 

Furthermore, our design uses a MEMS mirror as the scanning optics, which is compact and low-power. MEMS scanning is faster than mechanical motors, without similar wear-and-tear, and this allows for multiple fovea or regions-of-interest in a scene. Additionally, MEMS mirrors are neither limited to coherent illumination, like phase arrays, nor constrained to specific light wavelengths, like photonics-based systems. 

To demonstrate depth sensing flexibility, we first train a deep neural network for depth completion and show that it delivers high quality scene geometry. We evaluate this with different sampling patterns, including those that are concentrated in a region of interest specified by vision-based control.

\textbf{Our contributions} include building a novel adaptive LIDAR (Fig. \ref{fig:setup}) that enables flexible deep depth completion (Fig. \ref{fig:dc_fov}). We also confirm that LIDAR foveation improves sensing in areas of interest (Table \ref{table:dc_fov}). We provide analysis of receiver optics characteristics, particularly the issue of small aperture created by the MEMS mirror (Table \ref{table:optics}). Finally, we present a working vision-based adaptive LIDAR, showing that real-time foveation is feasible (Fig. \ref{fig:video}).

\section{Related Work}

\noindent \textbf{Common depth modalities: } Many high-quality depth sensors exist today. In Table \ref{fig:tabqual} we show qualitative comparisons with these. Our sensor is the first proof-of-concept, real-time, adaptive LIDAR.

{\begin{table*}
\centering
\resizebox{\linewidth}{!}{%
\begin{tabular}{|c|c|c|c|c|}
\hline                    
Sensor & Technology & Outdoors & Textureless & Adaptive \\
\hline                  
ELP-960P2CAM               & Conventional Passive Stereo   & \chk & \X   & \X \\
Kinect v2                  & Time-of-Flight (LED)          & \X   & \chk & \X \\
Intel RealSense                  & Structured Light Stereo (LED) & \chk   & \chk & \X \\
Velodyne HDL-32E            & Time-of-Flight (Laser)       & \chk & \chk & \X \\
Resonance MEMS / Intel L515             & Time-of-Flight (Laser)        & \chk & \chk & \X \\
Robosense  RS-LiDAR-M1            & Solid State Time-of-Flight (Laser)        & \chk & \chk & \X \\ 
\hline
\hline
Programmable Light curtains & Adaptive Structured Light        & \chk & \chk & \chk \\ 
\textbf{Our sensor} & Adaptive LIDAR        & \chk & \chk & \chk \\
\hline
\end{tabular}
}
\caption{\textbf{Our Adaptive LIDAR vs. other common modalities.} We compare common depth modalities such as stereo \cite{lucas1981iterative}, Kinect \cite{newcombe2011kinectfusion}, Velodyne \cite{hall2011high}, Robosense solid state LIDAR and Resonance MEMS sensors \cite{flatley2015spacecube,stann2014integration,krastev2013mems} such as the Intel L515. Our work is closest to programmable light curtains for flexible, structured light reconstruction~\cite{bartels2019agile,wang2018programmable}. This paper is an alternate research direction with an adaptive LIDAR. } \label{fig:tabqual}
\end{table*}
}

\noindent \textbf{MEMS/Galvo mirrors for vision and graphics:} MEMS mirror modulation has been used for structured light~\cite{raskar1998office}, displays~\cite{jones2007rendering} and sensing~\cite{nayar2004programmable}. In contrast to these methods, we propose to use angular control to increase sampling in regions of interest as in \cite{tilmon2020foveacam}. While MEMS mirrors have been used in scanning LIDARs, such as from NASA and ARL \cite{flatley2015spacecube,stann2014integration,krastev2013mems}, \emph{these are run at resonance with no control, while we show adaptive MEMS-based sensing.} Such MEMS control has only been shown \cite{kasturi2016uav} in toy examples for highly reflective fiducials in both fast 3D tracking and VR applications \cite{milanovic2017fast,milanovic2011memseye}, whereas we show results on real scenes. \cite{sandner2015hybrid,chan2019long} show a mirror modulated 3D sensor with the potential for flexibility, but without leveraging guided networks, and we discuss the advantages of our novel receiver optics compared to these types of methods. Galvo mirrors are used with active illumination for
light-transport \cite{hawkins2005dual} and seeing around corners \cite{o2018confocal}. Our closest work is the use of light curtains for flexible, structured light reconstruction~\cite{bartels2019agile,wang2018programmable}. In contrast, ours is a MEMS-mirror driven LIDAR system with an additional capability of increasing resolution in some region of interest. In this sense, we are the first to extend adaptive control \cite{breivik2011motion,tasneem2018dirrectionally,chan2019long}, to LIDAR imaging of real dynamic scenes. 

\noindent \textbf{Adaptive Scanning Lidars:} Commercially available systems from AEye and Robosense are designed to improve lidar-rgb fusion for large systems such as autonomous cars. In contrast, our goal is to impact small autonomous systems and our choice of MEMS mirror modulation and our optical innovations track these goals.
\cite{yamamoto2018efficient} propose a progressive pedestrian scanning method using an actively scanned LIDAR, but results are shown in simulation rather than on a hardware platform.  \cite{tasneem2018dirrectionally} propose directionally controlling the LIDAR scan, but these adaptive results have been shown only for static scenes. In contrast, we show a real-time adaptive LIDAR that works for dynamic scenes.

\noindent \textbf{Guided and Unguided Depth Completion:} The impact of deep networks on upsampling and superresolution has been shown on images, disparity/depth maps, active sensor data etc.~\cite{battrawy2019lidar,chen2018estimating,mal2018sparse,lu2015sparse,uhrig2017sparsity,riegler2016atgv,hui16msgnet} with a benchmark on the KITTI depth completion dataset \cite{uhrig2017sparsity}. Upgrading from sparse depth samples has been shown \cite{van2019sparse}, and guided upsampling has been used as a proxy for sensor fusion such as the work that has recently been done for single-photon imagers \cite{lindell2018single} and flash lidar \cite{gruber2019gated2depth}. In contrast, we measure sparse low-power LIDAR depth measurements and we seek to flexibly change the sensor capture characteristics in order to leverage adaptive neural networks such as ~\cite{liu2019neural,adaptivelidarstanford}.

\section{Sensor design}

Fig. \ref{fig:setup} shows our sensor design, which consists of a small aperture, large depth-of-field color camera, optically co-located with a MEMS-modulated LIDAR sensor. If the camera has a FOV of $\omega_{cam}$ steradians and a resolution of $I$ pixels, then the average pixel support is $\omega_{pix} = (\omega_{cam} / {I})$. If the LIDAR laser's beam divergence is $\omega_{laser}$ steradians, then the potential acuity increase from LIDAR to camera is ($\omega_{laser} / \omega_{pix})$. Guided depth completion seeks to extract this potential, and a foveated LIDAR can leverage this capability.

Next, in Sect. \ref{sect:trans}, we discuss the MEMS-modulated transmitter optics that enable compact, low-power, fast and foveated controlled scans. The cost, however, is that MEMS mirrors act as a small aperture that reduces the received radiance, when compared to large mirrors such as galvos. In the following Sect. \ref{sect:recev} we model the receiver optical design space, comparing FOV, volume and received radiance.

\subsection{MEMS Mirror based Transmitter Optics}
\label{sect:trans}

The transmitter optics consist of the pulsed light source and MEMS mirror. The LIDAR's beam is steered by the mirror, whose azimuth and elevation are given by changes in control voltages over time, $(\theta(V(t)), \phi(V(t))$ over the MEMS mirror FOV $\omega_{mirror}$. Controlling the scan $V(t)$ can enable attending to a region-of-interest given by an adaptive algorithm. The challenge in transmitter optics is to provide a powerful, narrow laser with low beam divergence, given by

\begin{equation}
\omega_{laser} \approx \frac{M^2 \ \lambda}{w_o \ \pi}
\end{equation}

\noindent where $M$ is a measure of laser beam quality and $w_o$ refers to the radius at the beam waist, which we use a proxy for MEMS mirror size. Previous MEMS-mirror modulated LIDAR systems cover this design space, from high-quality erbium fiber lasers with near-Gaussian profiles, used by \cite{stann2014integration} where $M$ is almost unity, to low-cost edge-emitting diodes, such as \cite{tasneem2018dirrectionally} where $M \approx 300$ on the diode's major axis. Our setup follows the low-cost diode route, with an additional two-lens Keplerian telescope to reduce the beam waist to $6 mm$ and an iris to match the MEMS mirror aperture instead of an optical fiber~\cite{chan2019long}. 

\begin{figure}[t]
    \centering
    \includegraphics[width=\linewidth]{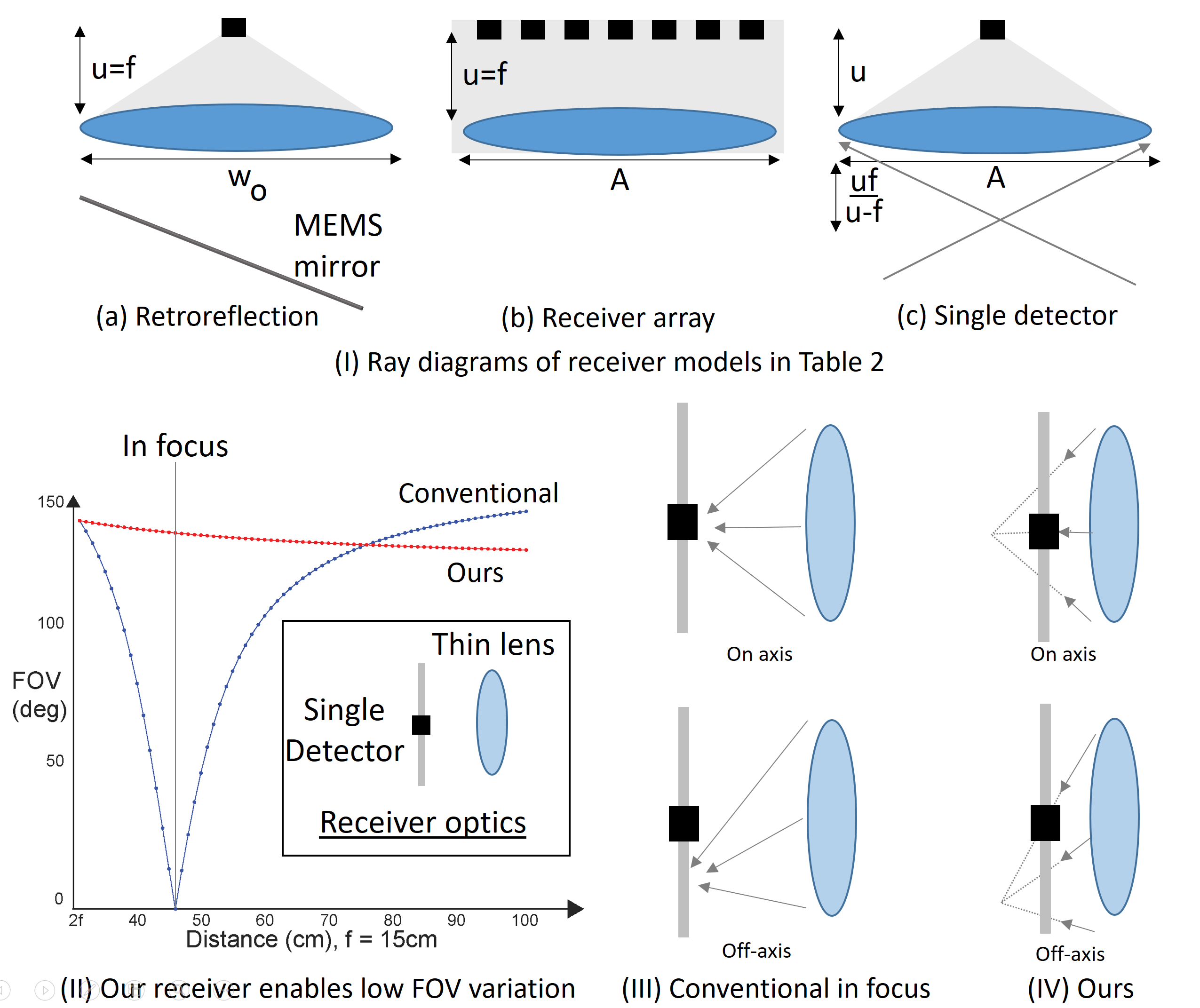}
    \caption{\textbf{Our proposed design vs. other designs.} In (I) we depict three common receiver designs, including retro-reflection (a), receiver array (b) and single detectors (c). Our design is a variant of (c), where we suggest a simple optical trick, such that the single detector is placed within the focal distance of the lens. This enables consistent FOV over range, as shown by the red curve in (II) and the designs in (III-IV). Simulations for a $f=15mm$ unit diameter lens.}
    \label{fig:trick}
\end{figure}
\subsection{Receiver Optics Design Tradeoffs}
\label{sect:recev}

{\begin{table*}\label{table:optics}
\centering
\resizebox{\linewidth}{!}{%
\begin{tabular}{|c||c|c|c|}
\hline                     
Design & Volume & FOV & Received Radiance \\ 
\hline                  
\hline 
Retroreflection 
& $\frac{\pi \ u \ w_o^2}{12}$
& = MEMS FOV $\omega_{\text{mirror}}$ 
& $\frac{atan(\frac{w_o}{2Z})}{\omega_{\text{laser}}Z tan(\frac{\omega_{\text{laser}}}{2})}$  \\
\hline 
Receiver array
& $u \ A^2$ 
& min$(2 \ atan(\frac{A}{2u})$, $\omega_{mirror})$
& $\frac{1}{2 \ Z \ tan(\frac{\omega_{laser}}{2})}$ \\
\hline
\textrm{Single detector} .
& \multirow{3}{*}{$\frac{\pi \ u \ A^2}{12}$}
& \multirow{3}{*}{min$(2atan(\frac{A(Z-f) \vert\vert \frac{Zu - fu-fZ}{Z-f} \vert\vert}{2ufZ})$, $\omega_{mirror}$)}
& \multirow{3}{*}{$\frac{1}{4Zatan(\frac{A(Z-f) \vert\vert \frac{Zu - fu-fZ}{Z-f} \vert\vert }{2ufZ}) tan(\frac{\omega_{\text{laser}}}{2})}$} \\
\textrm{\emph{Conventional}} \ $(u \geq f)$ &&& \\
\textbf{\emph{Ours}}  \ $(u < f)$ &&& \\
\hline
\end{tabular}} 
\vspace{-2mm}
\caption{\textbf{Receiver models.} Please see supplementary for derivations.}   \label{table:optics}
\end{table*}}

From the previous section, we can denote the transmitter optics design space as a combination of laser quality $M$ and MEMS mirror size $w_o$, which we write as $\Pi_t = \{M, w_o\}$. Now, we add receiver optics to the design space, which we denote as $\Pi_r = \{n,A,u,f\}$, with $n^2$ photodetectors in the receiver, $A$ as the aperture, $u$ as the distance between the photodetector array and the receiver optics, and $f$ as the focal length of the receiver optics. Therefore the full design space consists of both receiver/transmitter optics, $\mathbf{\Pi} = \{ \Pi_r, \Pi_t\}$.

We define the characterization of any instance within the design space $\mathbf{\Pi}$ as consisting of field-of-view $\Omega$ steradians, received radiance $s$ and volume $V$ denoted as $\Xi = \{ \Omega, s, V\}$. The range $Z$ is determined by the received radiance and the detector sensitivity. Computing these parameters depends on the design choices made, and we provide simulations comparing three designs: retro-reflective receivers~\cite{halterman2010velodyne} (Fig. \ref{fig:trick}I(a)), receiver arrays~\cite{chan2019long} (Fig. \ref{fig:trick}I(b)) and single-pixel detectors~\cite{tasneem2018dirrectionally} (Fig. \ref{fig:trick}I(c)).

\begin{figure*}[h]
    \centering
    \includegraphics[height = 0.45\linewidth, width=\linewidth]{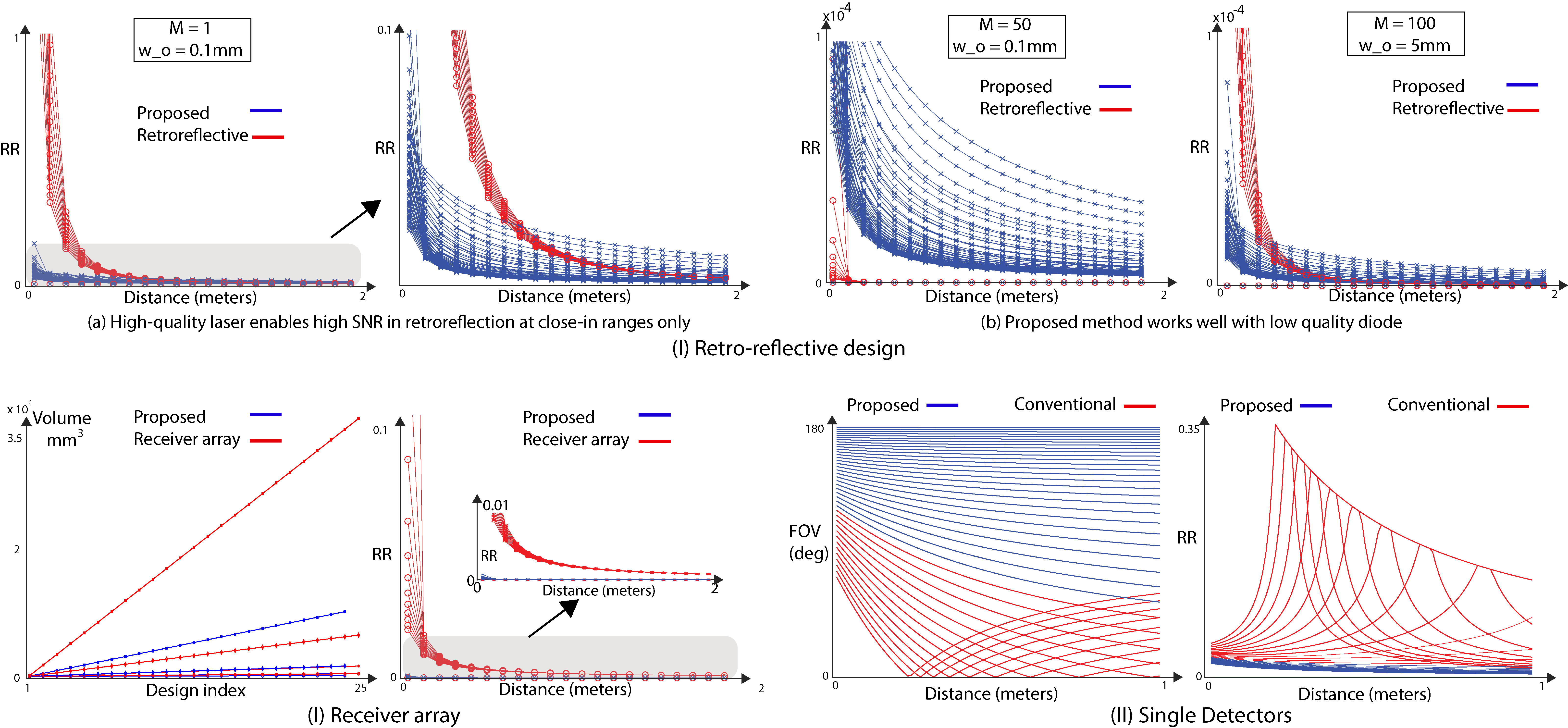}
    \caption{\textbf{Noiseless simulations comparing proposed method with other designs.} In (I) we compare the received radiance (RR) of proposed method with retroreflection for different laser qualities and mirror sizes. A high-quality laser (I)(a) enables higher RR for close-in scenes for retroreflective designs, but at large ranges, our method has higher RR. In (II)(a) we show that our proposed design has lower volume than a receiver array, across a wide range of focal lengths, but a receiver array has a higher RR (II)(b), even when compared to the best case for our sensor from (I). In (III) we compare our design with conventional single detectors, for a lens with $f=15mm$. Although our sensor shows consistent FOV ((III) left), it is always defocused, and faces an RR cost ((III) right).} \vspace{-0.05in}
    \label{fig:bigall}
\end{figure*}

\subsection{Simulation Setup and Conclusions}

Full derivations for the three receiver designs, shown in Table \ref{table:optics}, are in the supplementary material. The table refers to receiver sensor volume, field-of-view, and received radiance (normalized for a white Lambertian plane). The \emph{volume} is the convex hull of the opaque baffles that must contain the receiving transducer electronics and is either a cone or cuboid. The \emph{FOV} is the range of angles that the receiver is sensitive to, upper-bounded by the MEMS FOV $\omega_{mirror}$ \footnote{For simplicity, trignometric functions are written to act on steradian quantities, but in actuality act on the apex angle of the equivalent cone.}. \emph{Received radiance} is the area-solid angle product used in optics~\cite{mudge2019range} for a canonical LIDAR transducer, which can be loosely understood as loss of LIDAR laser dot intensity due to beam divergence and receiver aperture size when imaging a fronto-parallel, white Lambertian plane.  

In our noiseless simulations, we assumed a geometric model of light. To illustrate the trade-offs, we vary the laser quality between $M=1$ to $M=100$, representing an ideal Gaussian beam vs. a cheap laser diode. For the same reason, we vary the MEMS mirror size $w_o$ from $0.1mm$ (10 times larger than the TI DMD \cite{hornbeck1991architecture}) to $5mm$ (a large size for a swiveling MEMS mirror). The range of dimensions over which we explore the receiver design space are of the order of a small camera, with apertures $0cm \leq A \leq 10cm$, focal lengths $0mm \leq f \leq 50mm$ and image plane-lens distances $0mm \leq u \leq 50mm$. In Fig. \ref{fig:trick}(II)-(IV) we describe our proposed, simple modification to the conventional single-pixel receiver, where photodetector is placed on the optical axis, at a distance $v$ larger than the focal length $f$. In contrast to existing work on defocusing received radiances for FOV adjustment and amplitude compensation (e.g. \cite{mudge2019range}), we do not require special optics (e.g. split lens) and we have large off-axis FOV since the MEMS is not the aperture for the receiver. The \textbf{conclusion} from these experiments is that our sensor provides a new option for receiver design tradeoffs, explained next. 



\subsection{Design tradeoffs} 

For a full analysis of sensor design tradeoffs, please see the supplementary material. Here we compare our sensor vs. other designs. In the next section, we show how our proposed optical modification can improve single detectors even further.

\noindent \textbf{Retro-reflective receivers: } We show, in the supplementary, that retro-reflective designs are smaller than ours. The small retroreflective design also has the optimal FOV of the MEMS, due to co-location. Our design does have a received radiance advantage, since retroreflection requires the MEMS mirror to be the aperture for both receiver/transmitter. Fig. \ref{fig:bigall}(Ia) shows how this advantage eventually trumps other factors such as laser quality ($M=1$) or large mirrors. In the extreme case of low-cost diodes, Fig. \ref{fig:bigall}(Ib), our sensor has higher received radiance at close ranges too.

\noindent \textbf{Receivers arrays: } Large receiver arrays take up space, but they also have a higher received radiance, due to having a bigger effective aperture, when compared with our MEMS mirror. This is demonstrated in Fig. \ref{fig:bigall}(II) (right) for the particular case of $M=100,w_o=5mm$, favoring our design. Despite this, large arrays have higher received radiance at all depths. However, our design can always be made smaller, in every instance of these simulations. 

\noindent \textbf{Conventional Single detector: } Our approach is close to the conventional single pixel receiver, which can allow for detection over a non-degenerate FOV if it is defocused, as shown for a scanning LIDAR by \cite{tasneem2018dirrectionally}. When the laser dot is out of focus, some part of it activates the single photodetector. If the laser dot is in focus, the activation area available is smaller, but more concentrated. Next we describe and analyze our modification to the conventional single detector.

\subsection{Optical modification to single detectors}

Our approach is based on a simple observation; placing the image plane between the lens and the focus, i.e. $v<f$, will guarantee that the laser dot will never be in focus. For imaging photographs, this is not desirable, but for detecting the LIDAR system's received pulse, amplitude can be traded down, up to a point, as long as the peak pulse can be detected. Further, this optical setup ensures that the angular extent of the dot is nearly constant over a large set of ranges. This is further explained in the supplementary and supported by simulations (red curve) in Fig. \ref{fig:bigall}(III) (left) and explained in the ray diagrams of Fig. \ref{fig:bigall}(III) (right). 

For the conventional approach, when $u=f$, the FOV degenerates to a small value, where received radiance is also the highest. Our design does not suffer this depth-dependent FOV variation and is consistent across the range. As shown in the right, however, this results in a low received radiance since the system is always defocused. Simulations support this, in Fig. \ref{fig:bigall}(III), for $u>f$, shown in red, for settings of $f=15mm,A=100mm$, over a range of sensor sizes and ranges. In practice, we find consistent FOV to be more valuable than received radiance.

\section{Towards Adaptive LIDAR}



MEMS mirrors \cite{milanovic2017fast,milanovic2011memseye} are fast enough to enable adaptive sensing for dynamic scenes. The question then becomes \emph{how} to find good scan patterns, represented as voltage-dependent mirror angles over time, $(\theta(V(t)), \phi(V(t))$. These include open loop~\cite{breivik2011motion,ferrari2006track,tasneem2018dirrectionally,chan2019long} real-time estimation of regions of interest (ROIs) as well as end-to-end learning to decide where to sense next~\cite{liu2019neural,adaptivelidarstanford}. \textbf{Our contribution} here is to demonstrate LIDAR foveation for dynamic scenes with an open-loop algorithm based on motion detection ~\cite{ferrari2006track}.


\begin{figure*}
\centering
\includegraphics[width=1\linewidth]{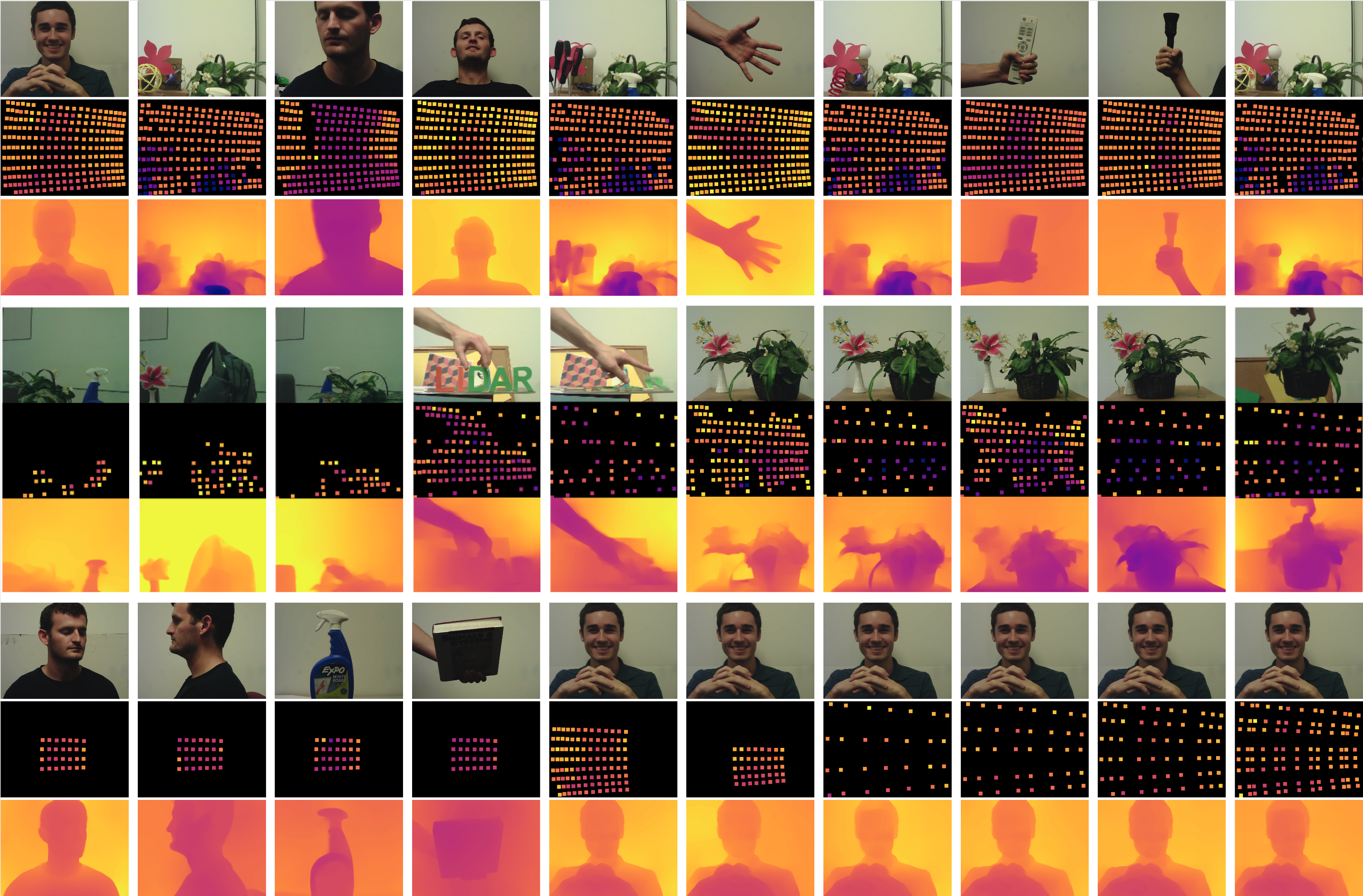} 
 \caption{\textbf{Adaptive Lidar Sampling.} This figure qualitatively demonstrates the flexibility of our adaptive LIDAR by showing a range of scan patterns. In row 1, a fixed, equi-angular full FOV scan pattern was used. In row 2, the density of the scan pattern was automatically adapted according to the RGB image's entropy. In row 3, columns 1-6, constant sampling density was applied on a rectangular ROI with maximal scene entropy. In row 3, columns 7-10, the FOV of the scan pattern was kept fixed and a sweep of the sampling density was performed. Note, with no depth samples, our depth completion model defaults to monocular depth estimation from the colocated camera, since we randomly sparsified the input depth maps during training to encourage robustness to a range of sampling densities (including zero samples).}
\label{fig:dc_fov}
\end{figure*}
\begin{table}[t]
\begin{center}
\centerline{
\tabcolsep=0.11cm
\begin{tabular}{c|ccc|ccc}
\toprule
\multirow{2}{*}{\textbf{Data}}  & \textbf{MRE} & \textbf{RMSE}  & \textbf{log}$\bm{_{10}}$ & $\bm{\delta_1}$ & $\bm{\delta_2}$ & $\bm{\delta_3}$ \\
                                         & (\%) & (m) & (m) & (\%)       & (\%)       & (\%)       \\
\hline 
Real & 10.16 & .1659 & .0410 & 89.80 & 95.88 & 98.63 \\
\bottomrule
\end{tabular}
}
\end{center}

\vspace{-2mm}
\caption{\textbf{LIDAR Evaluation.} The table reports the mean relative error (MRE), root mean squared error (RMSE), average ($\log_{10}$) error, and threshold accuracy ($\delta_i$) of the calibrated depth measurements, relative to the ``ground-truth'' Kinect V2 depths, over all 75 scenes of our real dataset. The Kinect V2 has an accuracy of 0.5\% of the measured range \cite{6964815}.}
\label{table:lidar_eval}
\end{table}

\noindent \textbf{Experimental setup: } Our LIDAR engine is a single beam Lightware SF30/C with an average power of $0.6mW$. This device can produce 1600 depth measurements per second at 100m. Data is captured as a stream of measurements, and each are time-stamped by the MEMS direction, given by the voltage $V(t)$. We modulate the single beam with a $3.6mm$ Mirrorcle MEMS mirror. Our current prototype has a range is $3m$ and a field-of-view of $\approx 25^\circ$. The laser dot, in steradians, is $6\times10^{-4} \Omega$ and this angular support is consistent over change in MEMS mirror angle. 

\begin{table}[t]
\begin{center}
\centerline{
\resizebox{1\columnwidth}{!}{
\tabcolsep=0.05cm
\begin{tabular}{c|c|ccc|ccc}
\toprule
\multirow{2}{*}{\textbf{Data}} & \multirow{2}{*}{\textbf{Method}} & \textbf{MRE} & \textbf{RMSE} & \textbf{log}$\bm{_{10}}$ & $\bm{\delta_1}$ & $\bm{\delta_2}$ & $\bm{\delta_3}$ \\
                      &             & (\%) & (m) & (m)  & (\%)       & (\%)       & (\%)       \\
\hline 
\multirow{2}{*}{NYU} &
Mono   & 8.55 & .3800 & .0361 & 90.56 & 98.08 & 98.56 \\
& Ours & \textbf{5.89} & \textbf{.2488} & \textbf{.0245} & \textbf{97.69} & \textbf{99.68} & \textbf{99.92} \\
\hline
\multirow{2}{*}{Real} &
Mono   & 28.26 & .3711 & .1090 & 50.14 & 87.38 & 96.00 \\
& Ours & \textbf{12.29} & \textbf{.1668} & \textbf{.0395} & \textbf{85.86} & \textbf{95.89} & \textbf{99.18} \\
\bottomrule
\end{tabular}
}
}
\end{center}
\vspace{-2mm}
\caption{\textbf{Base Comparison to Monocular Depth Estimation.} As a baseline, we compare to state-of-the-art monocular depth estimation \cite{alhashim2018high} (Mono) to our depth completion method (Ours) on a sub-sampled version of the NYUv2 Depth \cite{Silberman:ECCV12} (NYU) dataset and on our real dataset (Real). Both the monocular depth estimation and depth completion methods were trained only on NYUv2 data. To account for this, monocular depth estimates  were scaled by the ground-truth median, as in \cite{alhashim2018high}. Such scaling was not performed for depth completion predictions since the sparse LIDAR samples provide a reference absolute depth. }
\label{table:mono}
\end{table}

\noindent \textbf{Calibration and validation: } Since our sensor response is linear, we apply a 1D calibration to convert the LIDAR voltages into distances. We evaluate the quality of our sensor measurements and our calibration by computing the mean relative error (MRE), root mean squared error (RMSE), average ($\log_{10}$) error, and threshold accuracy ($\delta_i$) of the calibrated depth measurements from our LIDAR. We do this relative to the ``ground-truth'' Kinect V2 depths, over all 75 scenes of our real dataset, and these are reported in Table \ref{table:lidar_eval}. The Kinect V2 has an accuracy of 0.5\% of the measured range \cite{6964815}. Finally, we also captured 10 fronto-planar scenes (at ranges .5m-3m) and computed the RMSE of the depth measurements along the plane using the SVD method: the resulting average RSME was 0.06918m.


\subsection{Depth completion}
\label{sec:dc}

We now describe depth completion for the foveated measurements of our LIDAR. This builds on existing work~\cite{uhrig2017sparsity,zhang2018deep} where the sparse depth measurements are captured by our flexible LIDAR sensor and the ``guide'' image is captured by a RGB camera that is co-located with the sensor. We train a DenseNet-inspired \cite{huang2017densely} encoder-decoder network to perform RGB-guided depth completion of sparse measurements.

\noindent \textbf{Architecture.} We adopt \cite{alhashim2018high}'s encoder-decoder network architecture, except that our network has $4$ input channels, as it expects a sparse depth map concatenated with an RGB image. The encoder component of our network is the same as DenseNet 169 minus the classification layer. The decoder component consists of a three convolutional blocks followed by a final $3 \times 3$ convolutional layer. Each bilinear upsampling block consists of two $3 \times 3$ convolutional layers (with a leaky ReLU), and $2 \times 2$ max-pooling. 

\begin{table}[t]
\begin{center}
\centerline{
\resizebox{1\columnwidth}{!}{
\tabcolsep=0.11cm
\begin{tabular}{c|c|ccc|ccc}
\toprule
\multirow{2}{*}{\textbf{Data}} & \multirow{2}{*}{\textbf{FPS}} & \textbf{MRE} & \textbf{RMSE}  & \textbf{log}$\bm{_{10}}$ & $\bm{\delta_1}$ & $\bm{\delta_2}$ & $\bm{\delta_3}$ \\
                      &                      & (\%) & (m) & (m) & (\%)       & (\%)       & (\%)       \\
\hline 
\multirow{5}{*}{NYU} &
30       & 5.89 & .2488 & .0245 & 97.69 & 99.68 & 99.92 \\
& 24     & 5.88 & .2430 & .0244 & 97.97 & 99.70 & 99.92 \\
& 18     & 5.59 & .2261 & .0233 & 98.50 & 99.77 & 99.94 \\
& 12     & 5.65 & .2255 & .0236 & 98.52 & 99.77 & 99.94 \\
& 6      & 5.15 & .1879 & .0217 & 99.32 & 99.91 & 99.98 \\
\hline
\multirow{5}{*}{Real} &
30       & 12.29 & .1668 & .0395 & 85.86 & 95.89 & 99.18 \\
& 24     & 12.09 & .1644 & .0446 & 86.34 & 96.04 & 99.26 \\
& 18     & 11.57 & .1578 & .0430 & 87.27 & 96.61 & 99.30 \\
& 12     & 11.59 & .1558 & .0435 & 88.26 & 97.01 & 99.33 \\
& 6      & 11.19 & .1537 & .0422 & 88.10 & 97.19 & 99.26 \\
\bottomrule
\end{tabular}
}
}
\end{center}
\vspace{-2mm}
\caption{\textbf{Evaluation of Depth Completion.} This table conveys three key features of our system: (1) It highlights, the trade-off between frame rate and depth uncertainty, which impacts real-time applications; (2) it provides a quantitative evaluation of the robustness of our depth completion algorithm to varying sampling densities; and (3) provides an illustrative example of our system flexibility, which can be leveraged for a range of applications. For frame rates of 30, 24, 18, 12 and 6, the samples per frame were 28, 40, 60, 104 and 231 respectively. }
\label{table:dc_sub}
\end{table}

\begin{table}[t]
\begin{center}
\centerline{
\resizebox{1\columnwidth}{!}{
\tabcolsep=0.05cm
\begin{tabular}{c|c|ccc|ccc}
\toprule
\multirow{2}{*}{\textbf{Data}} & \multirow{2}{*}{\textbf{Method}} & \textbf{MRE} & \textbf{RMSE} & \textbf{log}$\bm{_{10}}$ & $\bm{\delta_1}$ & $\bm{\delta_2}$ & $\bm{\delta_3}$ \\
                      &                      & (\%) & (m) & (m)  & (\%)       & (\%)       & (\%)       \\
\hline 
\multirow{2}{*}{NYU} &
Full FOV   & 5.52 & .2392 & .0231 & 98.24 & 99.86 & 99.98 \\
& Foveated & \textbf{4.81} & \textbf{.1845} & \textbf{.0202} & \textbf{99.50} & \textbf{99.97} & \textbf{100} \\
\hline
\multirow{2}{*}{Real} &
Full FOV   & 15.72 & 0.1925 & .0566 & 80.30 & 99.79 & 99.36 \\
& Foveated & \textbf{13.36} & \textbf{.1589} & \textbf{.0497} & \textbf{83.24} & \textbf{97.80} & \textbf{99.46} \\
\bottomrule
\end{tabular}
}
}
\end{center}
\vspace{-2mm}
\caption{\textbf{Depth Completion on Foveated Lidar Data}. ``Foveated'' means that the scan pattern was automatically adapted to densely sample a region of interest in the scene. ``Full FOV" means that a scene independent equi-angular scanning pattern was utilized. In all cases, the ``Foveated'' and ``Full FOV'' scan patterns contain the same number of samples (hence, the equivalent frame rates). Results are evaluated at 30 FPS. Both Full FOV and Foveated errors are computed only in identical regions of interest, showing foveation increases accuracy. }
\label{table:dc_fov}
\end{table}

\begin{figure}[t]
\centering
\includegraphics[width=1\linewidth]{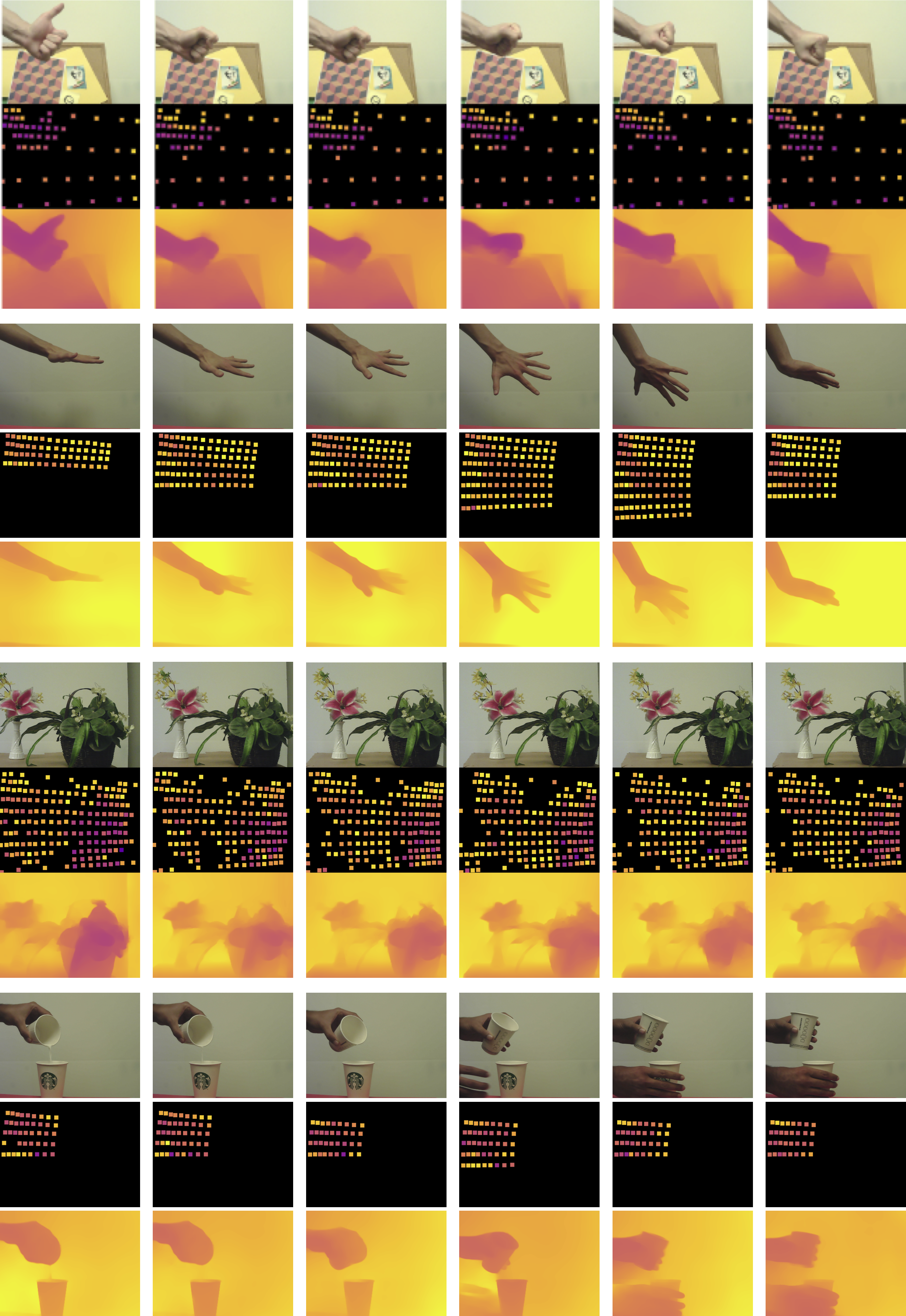}
\vspace{.5mm}
 \caption{\textbf{Motion-based adaptive sensing.} As the object moves, we use background subtraction to detect the region of interest and the MEMS-modulated LIDAR puts the samples where the object is located. Please see supplementary video.\vspace{-0.15in} }
\label{fig:video}
\end{figure}

\noindent \textbf{Optimization.}
We adopt \cite{alhashim2018high}'s loss as a weighted sum of three terms: 
\begin{equation}
L(y,\hat{y}) = \lambda L_{depth}(y,\hat{y}) + L_{grad}(y,\hat{y}) + L_{SSIM}(y,\hat{y})    
\end{equation}
where $y$ and $\hat{y}$ denote the ground-truth and estimated depth maps respectively, and $\lambda$ denotes a weighting parameter, which we set to $0.1$. The remaining terms are defined as in \cite{alhashim2018high} which has the full expressions.

\noindent \textbf{Datasets and Implementation: } We perform our evaluations using two datasets: a real dataset captured with our LIDAR system and a simulated Flexible LIDAR dataset generated by sub-sampling the NYUv2 Depth dataset \cite{Silberman:ECCV12}. The real dataset consists of pairs of RGB images and sparse depth measurements of 75 different scenes captured with our LIDAR system. For each of the 75 scenes, we also capture a dense ``ground-truth'' depth map using a Kinect V2 depth sensor that is stereo calibrated with our LIDAR system. All real dataset images are used exclusively for testing. The simulated dataset is split into non-overlapping train, test, and validation scenes.

We train the model described in section \ref{sec:dc} on a simulated Flexible LIDAR dataset generated by sub-sampling the NYUv2 Depth dataset. During training, depths were randomly scaled to prevent the network from overfitting to the color camera used to capture the RGB images in the NYUv2 dataset. For optimization, we used Adam \cite{kingma2014adam} with $\beta_1 = 0.9$, $\beta_2 = 0.999$, $\epsilon=1\text{e}-8$, a learning rate of $0.0001$, and a batch size of $4$. The learning rate was dropped to $0.00001$ after $94k$ iterations. The first layer used Xavier initialization, whereas other layers were initialized with the pre-trained weights from $\cite{alhashim2018high}$ for monocular depth estimation on NYUv2. To augment the data, we first randomly resize the input such that the smallest dimension varies between $640, 832,$ or $1024$. We then apply a random crop to the reduce the size to $640 \times 480$. In addition, the RGB channels were randomly shuffled. 

As a `sanity check', we confirm that guided depth completion outperforms monocular depth estimation using a state-of-the-art network \cite{alhashim2018high} in Table \ref{table:mono}.

\subsection{Motion-based Foveated Depth Sampling}

Our flexible platform allows us to ask if foveated LIDAR sampling improves depth measurements. We evaluate our guided depth completion network on LIDAR data captured with two different sampling regimes, full field-of-view sampling and foveated sampling in regions of interest, at various frame rates.

Table \ref{table:dc_sub} shows our evaluation for full field-of-view sampling. Table \ref{table:dc_fov} demonstrates that foveation improves reconstruction in a region of interest, with qualitative results in Figures \ref{fig:dc_fov} and \ref{fig:video}.

We also perform foveated sampling in real-time, using an open-loop motion-based system to determine the scan patterns. For a dynamic scene, a foveating LIDAR can have fewer samples in the right places, decreasing latency and improving frame-rate. In Fig. \ref{fig:video}, we show objects moving across the scene. At each instance, the system performs background subtraction to segment a motion mask. This mask drives the LIDAR sampling, which has less points than a full dense scan would have, and therefore has higher sampling rate. In each result, ROI sampling density was identically dense, and the rest-of-the-scene density was different and sparser. The amortized frame rates for the real-time foveated sequence in rows 1, 2, 3 and 4 of Fig. \ref{fig:video} are 20 FPS, 13 FPS, 9 FPS and 24 FPS. Without foveation, dense sampling over the entire scene would result in a frame rate of 6FPS, which is much lower. Note that as the object changes position, the ROI changes and the LIDAR senses a different area. If temporal sampling is not the focus, then the method can instead densely sample the points onto the region of interest, increasing the angular resolution (i.e. zooming).  Finally, we note that all results include depth completion of the measurements, showing high-quality results. 

\section{Discussion}

\noindent \textbf{Limitations:} Our LIDAR engine has a $3m$ range, which enables initial feasibility tests and is appropriate for certain tasks such as gesture recognition. Range extension is achievable, since the Lightware LIDAR electronics engine has a 100m outdoor range and is only reduced by unnecessary optical losses. For future prototypes we wish to remove these losses with a GRIN lens, as done by \cite{flatley2015spacecube}. 

\noindent \textbf{Conclusions:} Our LIDAR prototype enables the kind of adaptive sensing, which, thus far, has only been shown in simulation and follows a recent trend in computational photography to use data-driven approaches inside the sensor \cite{chakrabarti2016learning,chen2016asp,chang2018hybrid,sitzmann2018end}.

\bibliographystyle{ieee}
\bibliography{main.bib}

\newpage
\clearpage

\setcounter{page}{1}
\setcounter{section}{0}
\setcounter{figure}{0}
\setcounter{table}{0}
\setcounter{equation}{0}
\setcounter{footnote}{0}
\renewcommand{\thepage}{A\arabic{page}}
\renewcommand{\thesection}{\Alph{section}}
\renewcommand{\thefigure}{A\arabic{figure}}
\renewcommand{\thetable}{A\arabic{table}}
\renewcommand{\theequation}{A\arabic{equation}}

\def\httilde{\mbox{\tt\raisebox{-.5ex}{\symbol{126}}}}

\makeatletter
\def\@thanks{}
\makeatother

\pagenumbering{gobble}

\title{\vspace*{-2.5ex} Supplementary Material: \\ Towards a MEMS-based Adaptive LIDAR \vspace*{-2.0ex}}

\author{Francesco Pittaluga\thanks{Equal Contribution}\\
NEC Labs America\\
{\tt\small francescopittaluga@nec-labs.com}
\and
Zaid Tasneem$^{*}$\\
University of Florida\\
{\tt\small ztasneem@ufl.edu}
\and
Justin Folden$^{*}$\\
University of Florida\\
{\tt\small jfolden@ufl.edu}
\and
Brevin Tilmon\\
University of Florida\\
{\tt\small btilmon@.ufl.edu}
\and
Ayan Chakrabarti\\
Washington University in St. Louis\\
{\tt\small ayan@wustl.edu}
\and
Sanjeev J. Koppal\\
University of Florida\\
{\tt\small sjkoppal@ece.ufl.edu}
}

\maketitle

\section{Derivations}

Here we derive all the formulae in Table 1 for the three designs. We have provided the ray diagrams of the designs in Fig. \ref{fig:setup1} and we have reproduced Table \ref{fig:tabsupp} here.

\subsection{Volume}

For the retroreflection and single detector, the volume of the camera is a cone whose vertex is the location of the single detector. From the ray diagrams and from the equation of the volume of a cone, this is easily seen to be $\frac{\pi u w_o^2}{12}$ for the retroreflector and $\frac{\pi u A^2}{12}$ for the single detector. For the receiver array the volume is the entire enclosure, given by the volume of a cuboid, $u*A*A$.

\subsection{FOV}

The retroreceiver has the exact same FOV as the mirror, by definition. From \ref{fig:setup1}(b), the FOV of the receiver array is given by the vertex angle of the cone at the central pixel, given by $2atan(\frac{A}{2u})$, bounded by the FOV of the mirror. This assumes the receiver and transmitter are close enough to ignore angular overlap issues. 

{\begin{table*}
\centering
\resizebox{\linewidth}{!}{%
\begin{tabular}{|c|c|c|c|c|}
\hline                    
Sensor & Technology & Outdoors & Textureless & Adaptive \\
\hline                  
ELP-960P2CAM               & Conventional Passive Stereo   & \chk & \X   & \X \\
Kinect v2                  & Time-of-Flight (LED)          & \X   & \chk & \X \\
Intel RealSense                  & Structured Light Stereo (LED) & \chk   & \chk & \X \\
Velodyne HDL-32E            & Time-of-Flight (Laser)       & \chk & \chk & \X \\
Resonance MEMS / Intel L515             & Time-of-Flight (Laser)        & \chk & \chk & \X \\
Robosense  RS-LiDAR-M1            & Solid State Time-of-Flight (Laser)        & \chk & \chk & \X \\ 
\hline
\hline
Programmable Light curtains & Adaptive Structured Light        & \chk & \chk & \chk \\ 
\textbf{Our sensor} & Adaptive LIDAR        & \chk & \chk & \chk \\
\hline
\end{tabular}
}
\caption{\textbf{Our Adaptive LIDAR vs. other common modalities.} } \label{fig:tabsupp}
\end{table*}
}

To find the FOV of the single detector, consider the diagram in Fig. \ref{fig:setup3}, where the single detector is focused on the laser dot at distance $Z$ from the sensor. From similar triangles, the kernel size is given by first finding the in-focus plane at $u^{'}$ from the lens equation

\begin{equation}
\frac{1}{f} = \frac{1}{u^{'}} + \frac{1}{Z}
\end{equation}

\noindent and so $u^{'} = \frac{fZ}{(Z - f)}$. From the two vertex shared similar triangles on the left of the lens, we now have an expression for the kernel size 

\begin{equation}
kersize = abs(u - u^{'})*\frac{A}{u^{'}}
\end{equation}

\noindent Substituting the value of $u^{'}$, we get an expression for 
{
\begin{equation}
kersize = abs(u - \frac{fZ}{(Z - f)})*\frac{A}{\frac{fZ}{(Z - f)}}
\end{equation}

\begin{equation}
= \frac{A(Z-f)\|(Zu - fu-fZ)\|}{fZ\|Z-f\|} 
\end{equation}

\begin{equation}
= \frac{A(Z-f)\|\frac{Zu - fu-fZ}{Z-f}\|}{fZ}.
\end{equation}
}

\normalsize

\noindent From the figure, the FOV, given by kernelangle is 

\begin{equation}
2atan(\frac{kersize}{2u}) = 2atan(\frac{A(Z-f)\|\frac{Zu - fu-fZ}{Z-f}\|}{2ufZ})
\end{equation}

\subsection{Received Radiance (RR)}

From Fig. \ref{fig:setup2}, the power from the laser decreases with distance. This is just fall-off from the source, and we represent it here as the area of the laser dot on a fronto-parallel plane. From the figure, this can be calculated from simple trignometry as 

\begin{equation}
2 Ztan(\frac{\omega_{laser}}{2}), 
\end{equation}

and we use the reciprocal for the RR as 

\begin{equation}
\frac{1}{2 Z tan(\frac{\omega_{laser}}{2})}.
\end{equation}

\noindent \textbf{1. Receiver array:} The receiver array is assumed to capture all the available radiance from the laser dot, and so the RR is exactly the same as the power fall-off described above as

\begin{equation}
\frac{1}{2 Z tan(\frac{\omega_{laser}}{2})}. 
\end{equation}

\begin{figure}[h]
    \centering
    \includegraphics[width=1\linewidth]{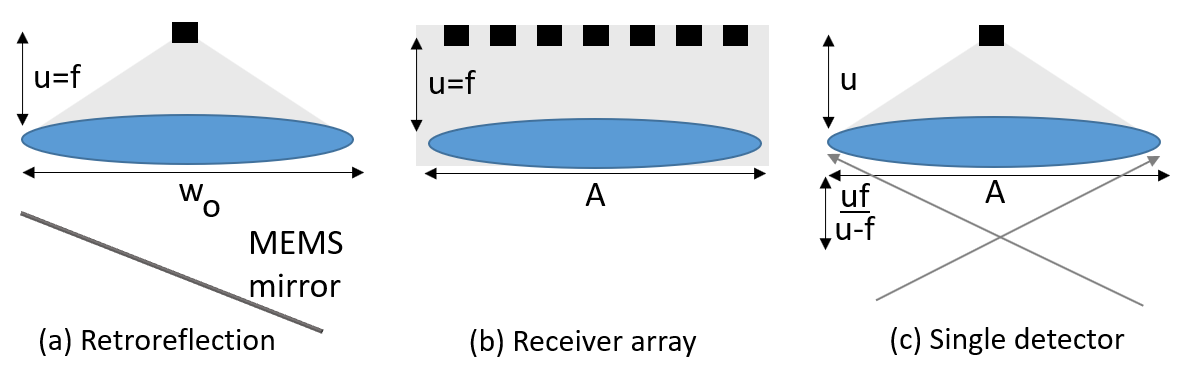}
    \caption{\textbf{Ray diagrams of designs}}
    \label{fig:setup1}
\end{figure}

\begin{figure}[h]
    \centering
    \includegraphics[width=1\linewidth]{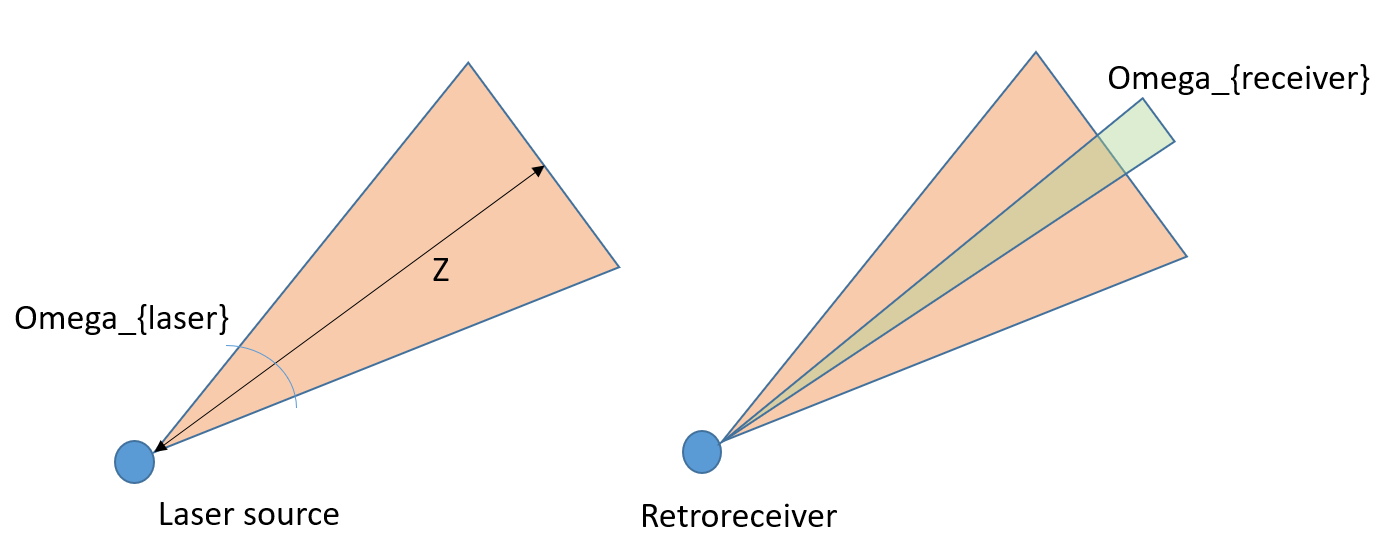}
    \caption{\textbf{Retroreflective Received Radiance}}
    \label{fig:setup2}
\end{figure}

\begin{figure}[h]
    \centering
    \includegraphics[width=1\linewidth]{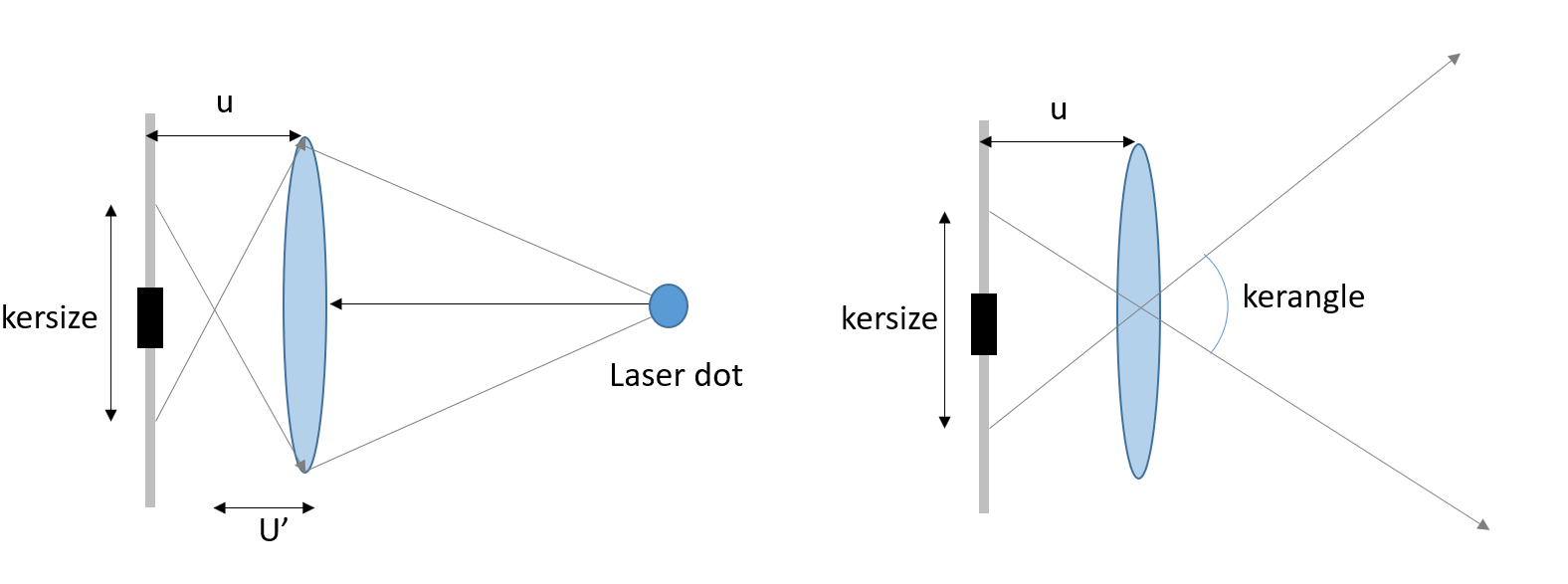}
    \caption{\textbf{Single Detector Received Radiance}}
    \label{fig:setup3}
\end{figure}

\noindent \textbf{2. Retroreflection:} As can be seen in the right of  Fig. \ref{fig:setup2}, the ratio of the received angle to the transmitted angle gives the fraction of the received  radiance from the laser dot. From Fig. \ref{fig:setup1}(a), the single detector receives parallel light of width $w_o$. For any particular depth $Z$ therefore, the angle subtended by this width at the sensor decreases and is given by 

\begin{equation}
\omega_{receiver} = 2atan(\frac{w_o}{2Z}) 
\end{equation}

and the fraction of the fall-off received is given by

\begin{equation}
\frac{2atan(\frac{w_o}{2Z})}{\omega_{laser}}. 
\end{equation}

Multiplying this with the fall-off above gives

\begin{equation}
\frac{2atan(\frac{w_o}{2Z})}{\omega_{laser}} * \frac{1}{2 Z tan(\frac{\omega_{laser}}{2})} 
\end{equation}

\begin{equation}
= \frac{atan(\frac{w_o}{2Z})}{w_{laser}Ztan{\frac{\omega_{laser}}{2}}}.
\end{equation}

Note this assumes that
$\omega_{receiver}<\omega_{laser}$, and if this is not the case then a max function must be added so that $\omega_{receiver}$ does not exceed $\omega_{laser}$. \\

\noindent \textbf{3. Single detector:} We just reduce the fall-off by the ker-angle calculated before, and therefore the RR is 

\begin{equation}
\frac{1}{kerangle*2 Z tan(\frac{\omega_{laser}}{2})}
\end{equation}

\begin{equation}
= \frac{1}{4Zatan(\frac{A(Z-f)\|\frac{Zu - fu-fZ}{Z-f}\|}{2ufZ}) tan(\frac{\omega_{laser}}{2})}
\end{equation}

\section{Single detector and proposed modification}

Our approach is based on a simple observation; placing the image plane between the lens and the focus, i.e. $v<f$, will guarantee that the laser dot will never be in focus. For imaging photographs, this is not desirable, but for detecting the LIDAR system's received pulse, amplitude is less important than timing information (i.e. pulse peak in our case). Further, this optical setup ensures that the angular extent of the dot is nearly constant over a large set of ranges. To see this, consider the second column of the table for our design. When $u<f$ and $z >> f$, the FOV becomes $2atan(\frac{A(f - u)}{2uf})$. Suppose $u << f$, then we can rewrite as $2atan(\frac{A(1 - \frac{u}{f})}{2u})$, which becomes $2atan(\frac{A}{2u})$, which is near-constant. This is supported by simulations discussed in the main paper in Figures 2 and 3. 

Of course, for the conventional approach, when $u=f$, there is a low FOV since the laser dot is sharply in focus. This is supported by simulations in Figures 2 and 3 in the main paper, for settings of $f=15mm,A=100mm$, over a range of sensor sizes and ranges. Therefore, the FOV degenerates to a small value, where received radiance is also the highest. Our design does not suffer this depth-dependent FOV variation and is consistent across the range. However, as shown in the right of the figure, this results in a low received radiance since the system is always defocused. In practice we find the consistent FOV to be more valuable than received radiance, and, further, depth completion can improve raw measurements.

\subsection{Analysis of Sensor Design Tradeoffs} 

\noindent \textbf{Retro-reflective receivers: } If high-quality lasers such as erbium fiber lasers \cite{stann2014integration} are used, where $M$ is near-unity, then these can be coupled with a co-located receiver and a beamsplitter, as shown in Fig. \ref{fig:trick}I(a), where the detector lens distance is equal to the focal length $u=f$. Consider the second column from Table \ref{table:optics}. The ratio of retro-reflective volume to our sensor's volume is $\frac{w_o}{A}$, which is usually less than one, since MEMS mirrors are small. 

In other words, retro-reflective designs are smaller than ours. The small retroreflective design also has the optimal FOV of the MEMS, due to co-location. Our design does have a received radiance advantage, since retroreflection requires the MEMS mirror to be the aperture for both receiver/transmitter. Fig. \ref{fig:bigall}(Ia) shows how this advantage eventually trumps other factors such as laser quality ($M=1$) or large mirrors. In the extreme case of low-cost diodes, Fig. \ref{fig:bigall}(Ib), our sensor has higher received radiance at close ranges too.

\noindent \textbf{Receivers arrays: } If cost and size are not issues, the receiver can be made large, such as a custom-built, large SPAD array~\cite{chan2019long} or a parabolic concentrator for $1.5mm$ detectors~\cite{stann2014integration}. Comparing such arrays' volume, in Table \ref{table:optics}'s second column, we can easily see the cuboid-cone ratio of $12/\pi$ favors our design, and is unsurprisingly shown in Fig. \ref{fig:bigall}(II) (left) across multiple focal lengths. 

On the other hand, it is clear that a large receiver array would have higher received radiance, due to having a bigger effective aperture, when compared with our MEMS mirror. This is demonstrated in Fig. \ref{fig:bigall}(II) (right) for the particular case of $M=100,w_o=5mm$, favoring our design. Despite this, large arrays have higher received radiance at all depths.

\noindent \textbf{Conventional Single detector: } Our approach is close to the conventional single pixel receiver, which can allow for detection over a non-degenerate FOV if it is defocused, as shown for a scanning LIDAR by \cite{tasneem2018dirrectionally}. When the laser dot is out of focus, some part of it activates the single photodetector. If the laser dot is in focus, the activation area available is smaller, but more concentrated. Next we describe and analyze our modification to the conventional single detector.

\end{document}